\documentclass[letterpaper]{article} 
\usepackage{aaai2027}  
\usepackage[hyphens]{url}  
\usepackage{graphicx} 
\usepackage{natbib}  
\usepackage{caption} 
\usepackage{algorithm}
\usepackage{algorithmic}
\usepackage{amssymb,amsfonts,amsmath}
\usepackage{epsfig}
\usepackage{mathrsfs}
\usepackage{textcomp}
\usepackage{bbding}
\usepackage{pifont}
\usepackage{wasysym}
\usepackage{subcaption}
\usepackage{xcolor,colortbl}
\usepackage{overpic}
\usepackage{microtype}
\usepackage{multirow}

\usepackage{newfloat}
\usepackage{listings}
\DeclareCaptionStyle{ruled}{labelfont=normalfont,labelsep=colon,strut=off} 
\floatstyle{ruled}
\newfloat{listing}{tb}{lst}{}
\floatname{listing}{Listing}

\usepackage{booktabs}
\newcommand{\jshi}[1]{{\color{black}#1}}
\newcommand{\js}[1]{{\color{black}#1}}
\title{C$^{2}$-INR: Customized Convolutional Implicit Neural Representation}
\author{
    Jinglei Shi\textsuperscript{\rm 1}\equalcontrib\corresponding,
    Xinran Chang\textsuperscript{\rm 1}\equalcontrib,
    Jiaqi Cui\textsuperscript{\rm 1},
    Yingjie Xia\textsuperscript{\rm 1},
    Zhaolin Xiao\textsuperscript{\rm 2},
    Chongyi Li\textsuperscript{\rm 1}
}
\affiliations{
    \textsuperscript{\rm 1}VCIP \& TMCC \& DISSec, College of Computer Science, Nankai University\\
    \textsuperscript{\rm 2}College of Computer Science and Engineering, Xi'an University of Technology\\
}

\begin{document}

\maketitle
\begin{abstract}
\label{sec:abs}
Implicit Neural Representation (INR) leverages neural networks to represent discrete signals such as images as continuous ones, where the network weights serve as a compact form of the signal itself.
Most existing INR methods adopt Multi-Layer Perceptrons (MLPs) as their backbone. \js{Since these models render each pixel independently, they inherently fail to exploit the spatial correlations that exist between neighboring pixels}.
In contrast, convolutional INRs can process pixels in parallel while inherently accounting for inter-pixel dependencies, making them a more natural fit for representing images.
Nevertheless, convolutional INRs remain relatively underexplored, and the majority of them rely on fixed architectural settings, leaving little room for image-specific adaptation.
In this paper, we investigate network customization for convolutional INRs. 
We replace conventional filters with irregular directional kernels, whose allocation is guided by the directional energy in the image spectrum, i.e., directions exhibiting stronger energy are assigned a larger number of kernels, enabling content-tailored convolution settings. 
These kernels are further reformulated via an orthogonal basis to achieve a superior sparse representation.
Moreover, we introduce an annealed Gumbel-Softmax-based mechanism for kernel-level activation function selection, which gives the most suitable activation function for each convolution kernel.
Extensive experiments demonstrate that our method, namely C$^{2}$-INR, achieves superior performance against state-of-the-art approaches under comparable parameter budgets across a wide range of image processing tasks, including representation, inpainting, and super-resolution. 
\end{abstract}

\section{Introduction}
\label{sec:intro}
INR methods achieve continuous representation of discrete signals by mapping input coordinates to target values such as pixels~\cite{mildenhall2021nerf,sitzmann2020implicit}, depth~\cite{park2019deepsdf}, or other attributes~\cite{srinivasan2021nerv} via an MLP. Their advantages include compact representation through network parameter fitting and greater flexibility in solving inverse problems like super-resolution and inpainting.
The performance of MLP-based INRs (M-INRs) largely depends on network architecture. To further boost their representation capability, recent work has explored various design strategies. For instance,~\cite{shi2024inductive} proposed an inductive gradient adjustment method that generalizes eNTK-based gradient transformation to mitigate spectral bias. Inspired by iterative refinement,~\cite{haider2026inr} introduced an iterative INR to strengthen high-frequency detail representation.~\cite{zhao2025adaptive} designed a high-frequency perception approach that uses the locations of high-frequency components as centers for wavelet positional encoding, balancing convergence across frequency bands. To address limited generalization,~\cite{li2023regularize} presented a Dirichlet-energy-based regularization that improves INR performance on non-uniform signals. Meanwhile,~\cite{zhang2026understanding} explored the role of bias terms and suggested initializing input-layer biases with pre-trained features to improve fitting quality. On activation functions,~\cite{heidari2024sl2a,han2026implicit} and~\cite{jayasundara2025mire} respectively proposed learnable single-layer activations, multi-scale sine activations, and layer-wise optimal activation matching, enhancing expressiveness from the activation perspective.

However, most existing INR research has centered on MLP-based backbones, which map coordinates to outputs (e.g., pixels) independently.
In contrast, convolution-based INRs (C-INRs) can better leverage spatial correlations via parallel convolution operations. Early C-INRs were introduced by~\cite{heckel2018deep}, which used uniform noise as input and cascaded convolutions as backbone.~\cite{chen2021nerv} extended this paradigm to video data, inspiring subsequent works~\cite{chen2023hnerv,zhang2024boosting,kwan2024immersive}. For light field compression,~\cite{shi2024learning} proposed a C-INR that splits kernels into spatial descriptors and angular modulators. 
Despite these applications, most existing C-INR methods rely on fixed network configurations (e.g., activation functions, kernel shapes, and channel numbers) across all scenes, which severely limits their representational capacity.

\begin{figure}[t]
    \centering
\includegraphics[width=0.49\textwidth,height=0.25\textwidth]{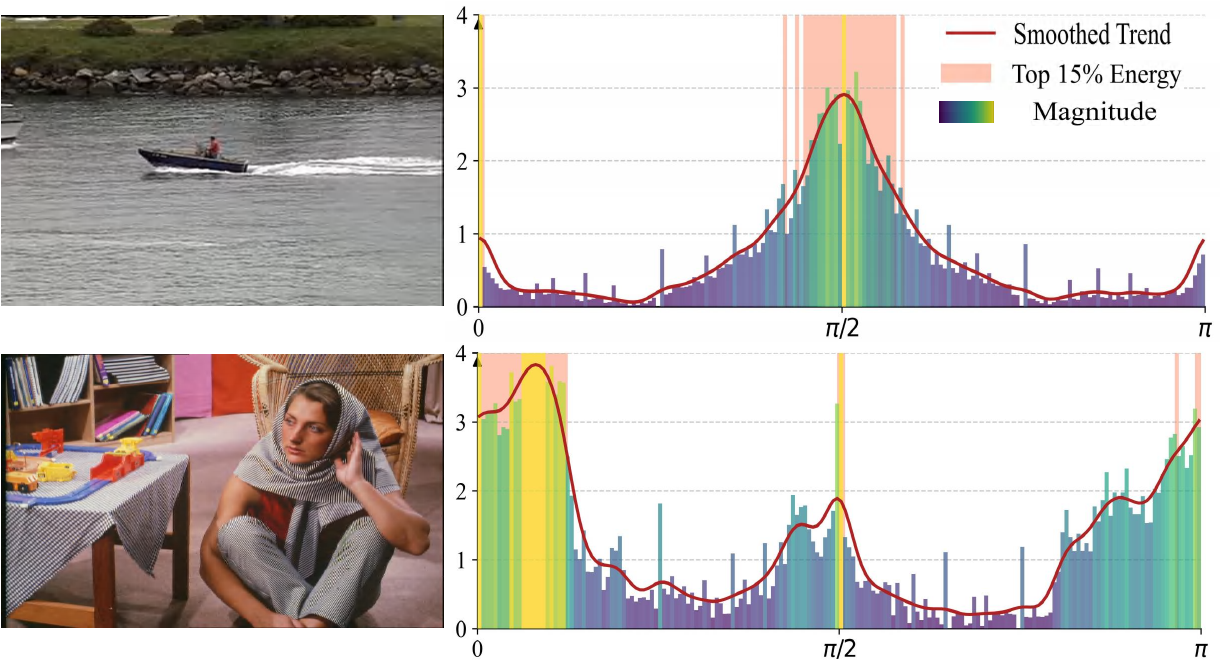}
    \caption{\js{Illustration of example images and their directional spectral magnitude distributions. The top image exhibits predominantly vertical variations, with its large-magnitude spectral components concentrated around $\pi/2$, while the bottom image contains mostly horizontal variations, with its energy peaking around 0. This implies that different images should have different optimal INR architectures.}}
    \label{fig:example_spectrum}
\end{figure}

\js{As shown in Fig.~\ref{fig:example_spectrum}, images with different content exhibit distinct characteristics (like spectrum distribution), suggesting that the optimal network architecture should also vary accordingly. This observation motivates our concept of network customization for C-INR, leading to the proposed C$^{2}$-INR method.}
Specifically, we target two core components of the network: convolution kernels and activation functions. To this end, we introduce a Spectral Guided Directional Convolution Block (SGDCB) for the backbone, which leverages the statistical energy distribution of the target image across directional frequency ranges to allocate kernel numbers, and further applies an orthogonal basis reformulation to boost representation performance. Complementarily, we propose an Annealed Gumbel-Softmax-based Activation Function Selection (AGS-AFS) mechanism that evaluates each candidate function from a preset dictionary via annealed scores, and assigns the most suitable one to each convolution kernel. Together, SGDCB and AGS-AFS yield a C-INR customized in terms of the target image. 
Experimental results demonstrate that our dual customization strategy achieves superior performance against State-Of-The-Art (SOTA) methods across multiple image processing tasks.
Our contributions can be summarized as follows:
\begin{itemize}
    \item \js{We pioneer the concept of C-INR customization, wherein the network architecture specifically tailored to the content of the target image. Such convolutional network design greatly enriches the available architectural choices for INR-based tasks.}
    \item We develop a frequency energy-guided directional kernel allocation method to enhance network adaptivity, coupled with an orthogonal-basis-based kernel reformulation that enhances network compactness.
    \item We design an annealed Gumbel-Softmax-based activation function selection mechanism, which estimates scores to assign the optimal function to each kernel, overcoming the expressiveness bottleneck of a single function.
\end{itemize}

\section{Related Work}
\label{sec:related_work}

\subsection{Implicit Neural Representation}
INR techniques parameterize target signals using neural networks. Based on the underlying network architecture, they fall into two categories: MLP-based (M-INR) and convolution-based (C-INR) approaches. M-INRs typically employ MLPs to learn a continuous mapping from coordinates to signal values~\cite{mildenhall2021nerf}, enabling arbitrary-resolution signal representation. In contrast, C-INRs adopt cascaded convolutional architectures to learn mappings from coordinates~\cite{chen2021nerv} or uniform noise~\cite{heckel2018deep} to target signals, supporting multi-point parallel rendering.
Recent advances have extended INR techniques to a broad spectrum of signal representation tasks, including audio~\cite{su2022inras}, images~\cite{saragadam2023wire,ramasinghe2022beyond,jayasundara2025pin}, video~\cite{yan2024ds,gao2025givic,guo2025metanerv}, light fields~\cite{shi2024learning}, and immersive video~\cite{wu2024tetrirf,zhu2025implicit}. To further boost representational capacity, especially for high-frequency signal components, extensive efforts have been devoted to improving various aspects of INR architectures. On activation functions, periodic~\cite{sitzmann2020implicit,liu2024finer,han2026implicit}, Gaussian~\cite{ramasinghe2022beyond}, wavelet-based~\cite{saragadam2023wire}, and other forms~\cite{jayasundara2025pin} have been proposed. On network architectures, multi-scale processing~\cite{liu2020multi,zhao2024pnerv} as well as multi-scale feature fusion and hierarchical encoding~\cite{zhu2025msnerv,jiang2025hiif} have been introduced. On the convergence front, iterative refinement strategies~\cite{haider2026inr} have also been explored. Collectively, these approaches have significantly elevated the performance ceiling of MLP-based INRs, effectively alleviating the spectral bias problem.

\subsection{Implicit Neural Network Customization}
As discussed above, extensive efforts have been devoted to overcoming spectral bias and improving representational capacity through novel activation functions, network architectures, or training strategies. However, due to the diversity of target signals, a fixed network setting often fails to adapt to individual signals, resulting in inherent performance limitations. Therefore, an important line of research focuses on customizing INRs to different targets. For example, some methods adjust activation function parameters via backpropagation during training~\cite{goyal2019learning,bingham2022discovering,fakhoury2022exsplinet}, or select optimal activations from a layer-wise dictionary~\cite{jayasundara2025mire}, to better capture complex patterns. \cite{tang2025canerv} designs an INR as an intelligent representation system that dynamically adapts its architecture to input content. \cite{ashkenazi2024towards} introduces the concept of editable INR and proposes a novel architecture that supports cropping operations. \cite{kazerouni2024incode} presents a harmonizer-composer framework for adaptively optimizing task-specific parameters in INR. Nevertheless, most existing customization efforts remain centered on MLP-based INRs, while customization for C-INR architectures remains largely underexplored.
\begin{figure*}[t]
    \centering
\includegraphics[width=0.999\textwidth]{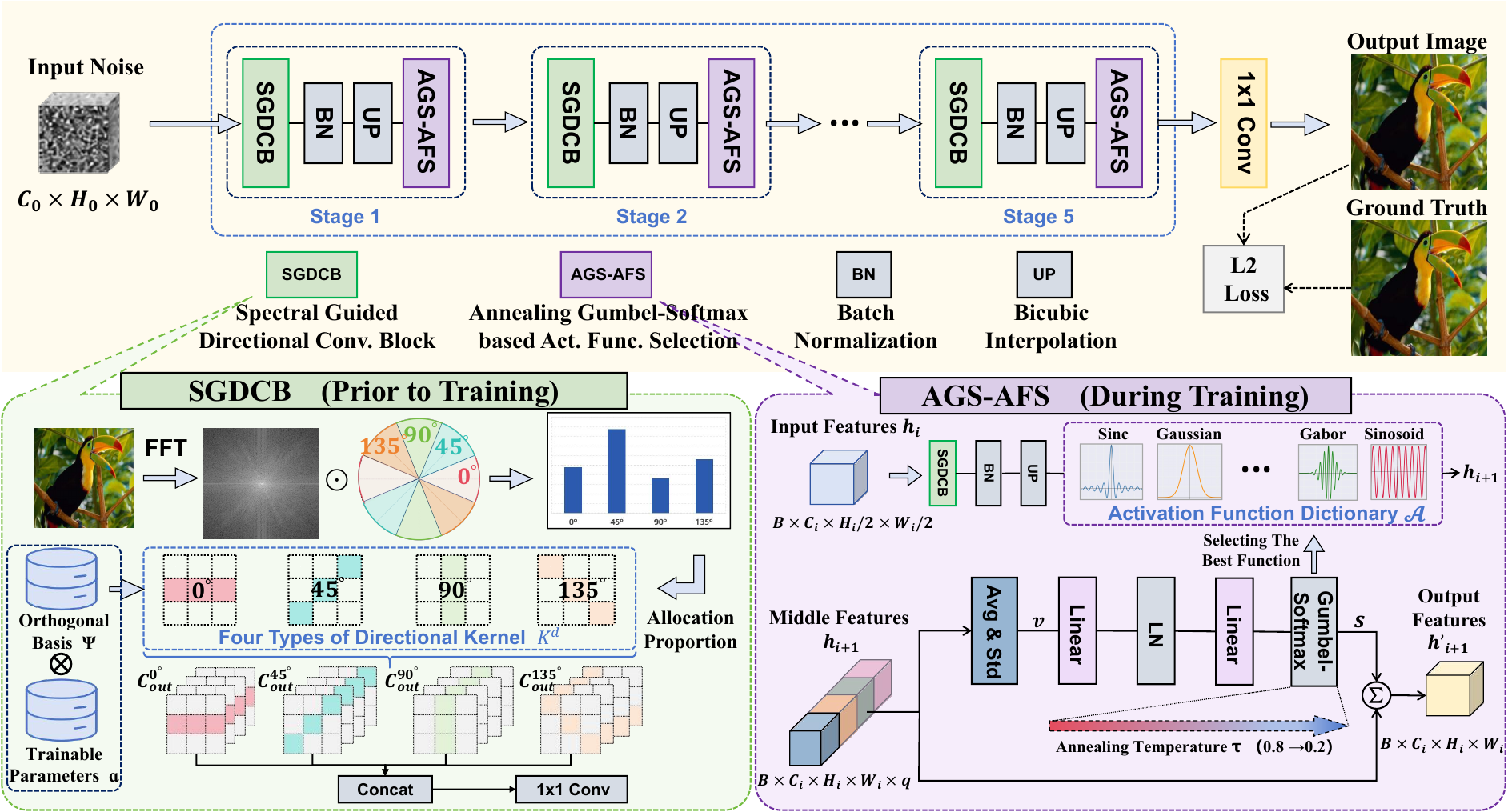}
    \caption{Overview of the proposed C$^{2}$-INR. The upper part illustrates the overall network backbone, which consists of five stages, each comprising SGDCB, BN, UP, and AGS-AFS modules, followed by a final $1\times 1$ convolutional decoding layer. The structure of SGDCB (bottom left) is determined prior to training: it focuses on analyzing the frequency energy in each direction and uses this information to allocate directional kernel numbers. These kernels are then reformulated via orthogonal basis decomposition with learnable parameters. Meanwhile, AGS-AFS (bottom right) operates during training, where it estimates scores for each candidate activation function using the Gumbel-Softmax function and adopts an annealing mechanism to gradually select the best-fitting activation function for each kernel.}
    \label{fig: main}
\end{figure*}

\section{Preliminary Knowledge}

Typical INRs use neural networks to construct a mapping $F_{\theta}(\cdot)$ from a low-dimensional input, such as coordinates $\mathbf{x} \in \mathbb{R}^{m}$, to the sampled values of the target signal (e.g., pixel intensity of an image) $\mathbf{y} \in \mathbb{R}^{n}$, thereby achieving continuous modeling of discrete signals:
\begin{equation}
    \mathbf{y} = F_{\theta}(\mathbf{x}),\ \ \mathbf{x} \in \mathbb{R}^{m},\mathbf{y} \in \mathbb{R}^{n},
\end{equation}
where $\theta$ denotes the network parameters. A MLP is commonly employed as the backbone, with each layer consisting of a linear mapping $f_{\theta_{i}}(\cdot)$ followed by an activation function $\sigma_{i}(\cdot)$:
\begin{equation}
    F_{\theta}(\cdot)=\sigma_{L}\circ f_{\theta_L} \circ \sigma_{L-1}\circ f_{\theta_{L-1}} \circ \cdots \circ \sigma_{1} \circ f_{\theta_{1}}(\cdot).
\label{eq:cascaded_equation}
\end{equation}
The network weights $\theta$ are optimized by minimizing the mean squared error between the predicted output $\mathbf{y} = F_{\theta}(\mathbf{x})$ and the ground truth $\mathbf{y}_{gt}$:
\begin{equation}
    \theta^{*} = \arg\min_{\theta}\left\|\mathbf{y}_{gt}-F_{\theta}(\mathbf{x})\right\|_2^2.
\end{equation}

Besides MLP, A CNN can also serve as the backbone for INR, giving rise to C-INRs. In this setting, uniform noise $\mathbf{x} = \epsilon \sim \mathcal{U}(0,1)$ is taken as input and multiple sampling points are rendered in parallel as the output $\mathbf{y}$. Accordingly, the parameters $\{\theta_{i}\}$ in Eq.~\ref{eq:cascaded_equation} correspond to convolutional kernel weights rather than those of fully connected layers.

\section{Methodology}
\label{sec:method}

\subsection{Network Backbone}
The goal of our work is to obtain a compact INR that not only enables image representation and other tasks but also maintains high rendering quality. As illustrated in Fig.~\ref{fig: main}, we adopt a five-layer cascaded convolutional network as our backbone to construct a C-INR. It takes a uniform noise tensor $\epsilon \sim \mathcal{U}(0,1)$ as input, with each layer sequentially performing convolution, batch normalization (BN), bicubic upsampling (UP), and activation function. The final output features are decoded via a $1\times 1$ convolutional layer and supervised by the target image $I_{gt}$.
However, employing predefined convolutions and activation functions leads to limited representational capacity when handling diverse target images. We thus focus on customizing these two components in terms of different images to improve the representation quality.

\subsection{Spectral Guided Directional Convolution Block}
Existing C-INRs~\cite{heckel2018deep,chen2021nerv,shi2024learning} rely on fixed, predefined convolutional blocks (typically with $3\times 3$ kernels). However, images of varying content often exhibit distinct characteristics: some scenes may be dominated by horizontal textures, while others contain predominantly diagonal ones. Applying identical kernel setting across such diverse scenes inevitably limits network performance.
Although early works~\cite{zhou2017oriented,weiler2018learning} have explored directional enhancements for convolutions, adapting such ideas to the INR setting is challenging. On one hand, image representation tasks demand high compactness, which complex kernel designs would compromise. On the other hand, it remains an open question how to extract effective cues from the input to guide kernel customization. Motivated by these considerations, we propose a customized building block for C-INRs, termed the Spectral Guided Directional Convolution Block (SGDCB), with the following key designs:

\subsubsection{Directional Kernel Allocation}
\label{sec:DKA}
Instead of the standard $3 \times 3$ convolution that covers all nine local pixels, we adopt slimmer directional convolutions that cover only three pixels along specific orientations. Specifically, these kernels operate in four directions: horizontal ($0^{\circ}$), vertical ($90^{\circ}$), main diagonal ($45^{\circ}$), and secondary diagonal ($135^{\circ}$). They can be efficiently implemented by applying shifting and rotation operations to $3 \times 1$ kernels. These four directional kernels are not only lightweight in terms of parameters but also offer stronger representational capacity along their orientations.
Images that exhibit stronger textural variations along a particular direction demand greater representational capacity in that direction. Accordingly, given the desired number of output channels $C_{out}$ for each layer, the number of kernels allocated to each direction should be reasonably determined based on the image content. 
To this end, we measure the energy of the target image $I_{gt}$ across different directional ranges in the frequency domain, using it as a cue for allocation:
\begin{equation}
    C^{d}_{out} = round( \frac{\sum_{\mathbf{p}\in I_{gt}} |M^{d-22.5^{\circ}}_ {d+22.5^{\circ}}\cdot\mathcal{F}(I_{gt})|}{\sum_{\mathbf{p}\in I_{gt}} |\mathcal{F}(I_{gt})|}\cdot \gamma C_{out}),
\label{eq:energy}
\end{equation}
where $\mathcal{F}(\cdot)$ denotes the Fourier Transform, $\mathbf{p}$ denotes each pixel in the target image $I_{gt}$, $d \in \{0^{\circ}, 45^{\circ}, 90^{\circ}, 135^{\circ}\}$ represents the orientation of the convolutional kernel, and $M_{d-22.5^{\circ}}^{d+22.5^{\circ}}$ is a binary mask that covers elements within a $45^{\circ}$ range centered at $d$ (like shown in bottom left of Fig.~\ref{fig: main}). $round(\cdot)$ is a rounding function, and $\gamma$ is a hyperparameter to adjust the number of parameter.
Given the constraint that $\sum_{d} = \gamma C_{out}$, Eq.~\ref{eq:energy} implies that a larger proportion of convolutional channels will be allocated to directions exhibiting stronger frequency energy, thereby enhancing the representational capacity.
Finally, the four directional branches are concatenated and fused with $\times 1 \times 1$ convolution kernels.

\subsubsection{Kernel Reformulation with Orthogonal Basis}
We also employ the closed-form solution of the discrete Chebyshev transform with a window length of $L_{w}=3$ to construct orthogonal basis $\Psi =[\psi^{0},\psi^{1},\psi^{2}]\in \mathbb{R}^{3\times 3\times 1}$, and then use them to reformulate the directional kernels. 
These three bases of $\Psi$ respectively captures smooth regions (DC component), edges (first-order differential) and texture patterns (second-order differential) characteristics. The directional convolution kernel $K^{d}\in \mathbb{R}^{C_{in}\times C^{d}_{out}\times 3\times 1}$ can be represented with the trainable coefficients $\alpha$ and the global orthogonal basis $\Psi$ as follows:
\begin{equation}
    K^{d} = \alpha \otimes \Psi,\ \alpha \in \mathbb{R}^{C_{in} \times C^{d}_{out}\times 3},\Psi\in \mathbb{R}^{3\times 3\times 1},
\end{equation}
where $\otimes$ denotes matrix cross product along the last dimension of $
\alpha$ and the first dimension of $\Psi$.
We will show in the section of ablation study that such reformulation improves performance with barely no increase in parametric cost.

It is worth noting that both directional kernel allocation and reformulation are carried out prior to the training, and thus incur no extra overhead to the training procedure.

\subsection{Kernel-Level Activation Function Selection}
As another core component of INR, different activation functions exhibit distinct properties in gradient flow, numerical stability, and nonlinear expressiveness.
While recent works have explored complex or learnable activation functions for M-INRs~\cite{jayasundara2025pin,han2026implicit}, the optimization in C-INRs remains largely unexplored. A naive solution is to adopt the layer-wise selection strategy from MIRE~\cite{jayasundara2025mire}, which picks the best function for each layer from a candidate set. However, this treats the layer as the smallest selection unit. In CNNs, kernels (each corresponding to a feature map) offer a finer granularity than layers, and different kernels often capture distinct feature characteristics. Motivated by this, we propose an Annealed Gumbel-Softmax-based Activation Function Selection (AGS-AFS) mechanism that selects the optimal activation function for each individual kernel:
\subsubsection{Annealed Activation Function Selection}
As shown in bottom right of Fig.~\ref{fig: main}, given a set of $q$ activation functions $\mathcal{A} = \{\text{Gaussian}(\cdot), \text{Sinusoid}(\cdot), \dots\}$, we sequentially pass the input feature $\mathbf{h}_{i}$ through the convolution operation $f_{K_{i}}$, BN, UP and each activation function in $\mathcal{A}$:
\begin{equation}
    \mathbf{h}_{i+1} = \mathcal{A}\circ \text{UP}\circ \text{BN}\circ f_{K_{i}}(\mathbf{h}_{i}),
\end{equation}
where $\mathbf{h}_{i}\in \mathbb{R}^{B \times C_{i}\times H_{i}/2\times W_{i}/2}$, and $\mathbf{h}_{i+1}\in \mathbb{R}^{B \times C_{i} \times H_{i} \times W_{i} \times q}$ denote the input and middle features of the $i$-th layer.
We then compute the mean and variance of each feature map, yielding a statistical vector $\mathbf{v}\in \mathbb{R}^{B\times C_{i}\times 2q}$ from $\mathbf{h}_{i+1}$.
Subsequently, linear layers $\text{Linear}(\cdot)$ and layer normalization $\text{LN}(\cdot)$ expand the statistical vector from $2q$ to $64$ dimensions, and then projected it back to $q$ dimensions. An annealed Gumbel-Softmax function is then used to predict a score distribution $\mathbf{s}\in \mathbb{R}^{B\times C_{i}\times q}$ over the $q$ activation candidates for each feature map:
\begin{equation}
    \mathbf{s} = \underset{\tau}{\text{Gumbel-Softmax}}(\underset{64\rightarrow q}{\text{Linear}}(\text{LN}(\underset{2q\rightarrow 64}{\text{Linear}}(\mathbf{v})))),
\end{equation}
where $\tau$ is the temperature hyperparameter of the Gumbel-Softmax. The final output features $\mathbf{h'_{i+1}}\in \mathbb{R}^{B\times C_{i}\times H_{i}\times W_{i}}$ is then computed as a weighted sum over $\mathbf{h_{i+1}}$:
\begin{equation}
\mathbf{h'_{i+1}}=\underset{q}{\sum}\ \mathbf{s'}\cdot \mathbf{h_{i+1}},
\label{eq:weighted_sum}
\end{equation}
where $\mathbf{s'}$ is $\mathbf{s}$ expanded along the $H_i$ and $W_i$ dimensions.
As $\tau$ is gradually annealed during training, $\mathbf{s}$ progressively becomes a \textit{\textbf{one-hot}} distribution. In this case, the weighted sum in Eq.~\ref{eq:weighted_sum} effectively reduces to selecting a single activation function per feature map, which corresponds to selecting the optimal function for each convolution kernel.

\subsubsection{Progressive Layer Fitting Strategy}
Although the above activation selection mechanism is applicable to all layers simultaneously, we adopt a progressive layer‑by‑layer fitting strategy to ensure training stability. Specifically, we begin with the first layer: the selection mechanism is applied to this layer, and an auxiliary decoding layer is used to decode its output features, with a downsampled version of $I_{gt}$ providing supervision. After a few iterations, the optimal activation functions for the kernels in this layer are determined and fixed. We then move to the next layer and repeat the process until all layers are processed. Finally, the entire network is fine‑tuned to full convergence.

Note that, although AGS-AFS employs learnable linear layers during selection, these layers are detached and discarded after the selection process, retaining only the chosen activation functions. Hence, no extra parameters are introduced to the final network. Moreover, unlike SGDCB, which is determined prior to training, AGS-AFS operates during training. The two mechanisms are thus independent of each other and can jointly contribute to the representation capability of C$^{2}$-INR. 

\section{Experiments}

\subsection{Implementation and Metrics}
Our C$^{2}$-INR is implemented with PyTorch framework and optimized using the Adam optimizer. We initialize temperature hyperparameter $\tau$ to 0.8 and gradually anneal it to 0.2 after 600 iterations for each layer.
We compare C$^{2}$-INR against recent SOTA methods, including M-INR approaches such as INCODE~\cite{kazerouni2024incode}, SL$^{2}$A-INR~\cite{heidari2024sl2a}, MIRE~\cite{jayasundara2025mire}, MSA~\cite{han2026implicit}, as well as the C-INR baseline Deep Decoder (DD)~\cite{heckel2018deep}. We adopt both distortion-based metrics PSNR, SSIM~\cite{wang2004image} and a perceptual metric LPIPS~\cite{zhang2018unreasonable} to evaluate performance.

\begin{table}[t]
	\setlength\tabcolsep{2pt}
	\centering
	\resizebox{0.47\textwidth}{!}{
		\begin{tabular}{l|cc|cc|cc}
			\hline
			\toprule 
			& \multicolumn{2}{c|}{\#Param$\sim$108k} 
			& \multicolumn{2}{c|}{\#Param$\sim$215k} 
            & \multicolumn{2}{c}{\#Param$\sim$437k} 
			\\ 
			\cmidrule{2-7}
			\multicolumn{1}{l|}{\multirow{-2}{*}{Methods}} 
			& PSNR$\uparrow$ & SSIM$\uparrow$&PSNR$\uparrow$
			
			& SSIM$\uparrow$ & PSNR$\uparrow$&SSIM$\uparrow$ 
			\\ 
			\hline\hline
			DD    & \underline{32.58}  & \underline{0.8796} & 
            33.99 & 0.9081 &
            35.48 & 0.9305\\

            INCODE    & 25.97  & 0.6423 & 
            30.95 & 0.8159 &
            35.59 & 0.9025\\

            SL$^{2}$A-INR    & 32.27  & 0.8348 & 
            \underline{36.23} & \underline{0.9104} &
            \underline{42.51} & \underline{0.9698}\\

            MIRE    & 31.77  & 0.8750 & 
            33.11 & 0.9018 &
            35.52 & 0.9305\\

            MSA    & 31.90  & 0.8523 & 
            34.62 & 0.9085 &
            39.03 & 0.9626\\
            
			\hline\hline
			Ours  & \textbf{39.38}  & \textbf{0.9571}  & \textbf{44.51}  & \textbf{0.9846}  & 
            \textbf{51.17}  & \textbf{0.9966}  \\
			\toprule
	\end{tabular}}
    \caption{Quantitative results for the image representation task using C$^{2}$-INR (Ours) and other SOTA methods. The best and second-best results are highlighted in \textbf{bold} and \underline{underlined}.}
   \label{tab:image_representation}
\end{table}

\subsection{Experimental Results}
\subsubsection{Image Representation}
\jshi{
For the image representation task, we evaluate each method under three parameter budgets: low ($\sim$108k params), medium ($\sim$215k params), and high ($\sim$437k params). Specifically, we configure C$^{2}$-INR with $C_{out}=\{32,64,128\}$ and $\gamma=2.6$ (the configuration details of other methods can be found in supplementary materials).
All models are fitted to the 24 full-resolution images from the Kodak dataset~\cite{kodakdataset} following their official training settings. Table~\ref{tab:image_representation} summarizes the averaged results across the three budget levels, where we observe that C$^{2}$-INR consistently outperforms other SOTA methods under all settings. It not only surpasses MLP-based alternatives, but also achieves substantially larger improvements over existing convolution-based counterpart DD.
We also visualize the reconstructed images in the first two rows of Fig.~\ref{fig:re_img}, readers are encouraged to zoom in for more details. Under the same parameter budget, C$^{2}$-INR is able to preserve subtle textures and details, which implies stronger representation capability and also suggests a promising application in compression tasks.
}

\begin{table}[t]
	\setlength\tabcolsep{2pt}
	\centering
	\resizebox{0.47\textwidth}{!}{
		\begin{tabular}{l|ccc|ccc}
			\hline
			\toprule 
			& \multicolumn{3}{c|}{$\times$2 SR} 
			& \multicolumn{3}{c}{$\times$4 SR} 
			\\ 
			\cmidrule{2-7}
			\multicolumn{1}{l|}{\multirow{-2}{*}{Methods}} 
			& PSNR$\uparrow$ & SSIM$\uparrow$&LPIPS$\downarrow$
			& PSNR$\uparrow$ & SSIM$\uparrow$&LPIPS$\downarrow$\\
			\hline
            \hline
			DD    & \underline{26.45}  & \underline{0.7488} & 
            0.3181 & \underline{24.25} &
            \underline{0.6650} & \underline{0.4241}\\

            INCODE    & 26.06  & 0.6764 & 
            \underline{0.2413} & 20.90 &
            0.4047 & 0.5324\\

            SL$^{2}$A-INR    & 25.68  & 0.7215 & 
            0.2540 & 22.15 &
            0.5478 & 0.5415\\

            MIRE    & 24.72  & 0.6308 & 
            0.3108 & 19.57 &
            0.3240 & 0.6117\\

            MSA    & 25.50  & 0.7274 & 
            0.2541 & 22.42 &
            0.5558 & 0.4416\\
            
			\hline\hline
			Ours  & \textbf{28.60}  & \textbf{0.8410}  & \textbf{0.1720}  & \textbf{24.48}  & 
            \textbf{0.6772}  & \textbf{0.3984}  \\
			\toprule
	\end{tabular}}
    \caption{Quantitative results for the image $\times 2$ and $\times 4$ SR tasks using C$^{2}$-INR (Ours) and other SOTA methods. The best and second-best results are highlighted in \textbf{bold} and \underline{underlined}.}
   \label{tab:image_sr}
\end{table}

\begin{table}[t]
	\setlength\tabcolsep{2pt}
	\centering
	\resizebox{0.47\textwidth}{!}{
		\begin{tabular}{l|ccc|ccc}
			\hline
			\toprule 
			& \multicolumn{3}{c|}{30\% Pixels Masked} 
			& \multicolumn{3}{c}{50\% Pixels Masked} 
			\\ 
			\cmidrule{2-7}
			\multicolumn{1}{l|}{\multirow{-2}{*}{Methods}} 
			& PSNR$\uparrow$ & SSIM$\uparrow$&LPIPS$\downarrow$
			
			& PSNR$\uparrow$ & SSIM$\uparrow$&LPIPS$\downarrow$\\
			\hline\hline
			DD    & \underline{35.83}  & \underline{0.9437} & 
            0.0369 & \underline{33.63} &
            \underline{0.9276} & \underline{0.0423}\\

            INCODE    & 34.35  & 0.8340 & 
            0.0453 & 32.15 &
            0.7981 & 0.0720\\

            SL$^{2}$A-INR    & 34.85  & 0.8477 & 
            0.0467 & 32.44 &
            0.8039 & 0.2813\\

            MIRE    & 35.53  & 0.9398 & 
            0.0359 & 31.49 &
            0.8850 & 0.0903\\

            MSA    & 34.16  & 0.9354 & 
            \underline{0.0293} & 31.02 &
            0.8870 & 0.0797\\
            
			\hline\hline
			Ours  & \textbf{37.09}  & \textbf{0.9610}  & \textbf{0.0098}  & \textbf{34.10}  & 
            \textbf{0.9327}  & \textbf{0.0257}  \\
			\toprule
	\end{tabular}}
    \caption{Quantitative results for the image inpainting tasks using C$^{2}$-INR (Ours) and other SOTA methods, with 30\% and 50\% pixels being randomly masked. The best and second-best results are highlighted in \textbf{bold} and \underline{underlined}.}
   \label{tab:image_inpainting}
\end{table}

\begin{figure*}[ht]
    \centering
    \setlength{
    \tabcolsep}{1pt}
    \centering
\resizebox{0.999\textwidth}{!}{
\begin{tabular}{ccccccc}
Kodak-08  & 40.15dB & 28.41dB & 26.74dB & 29.73dB & 29.21dB & 28.39dB\\
\includegraphics[width=0.156\linewidth]{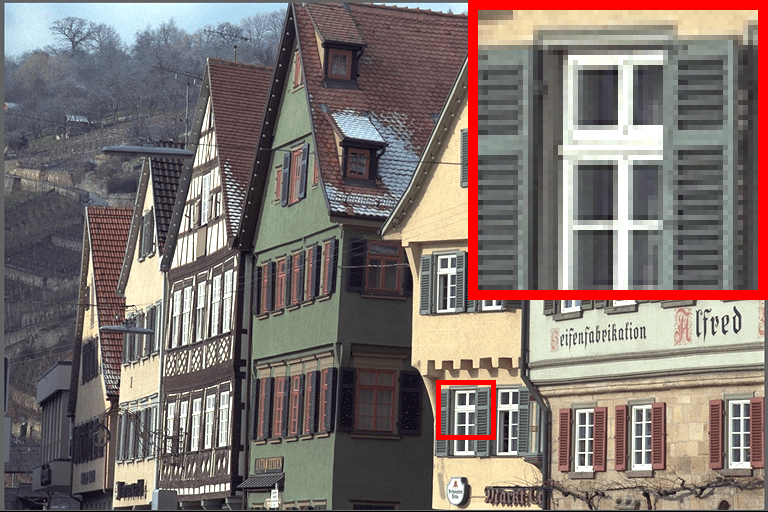} & 
\includegraphics[width=0.156\linewidth]{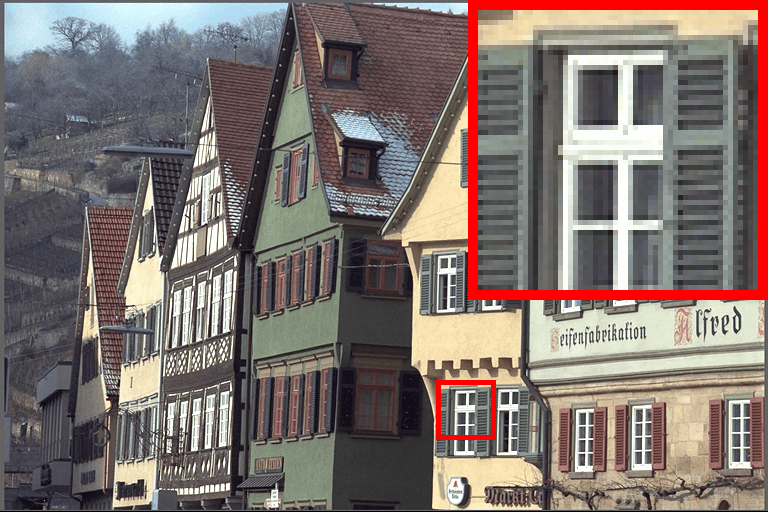} &
\includegraphics[width=0.156\linewidth]{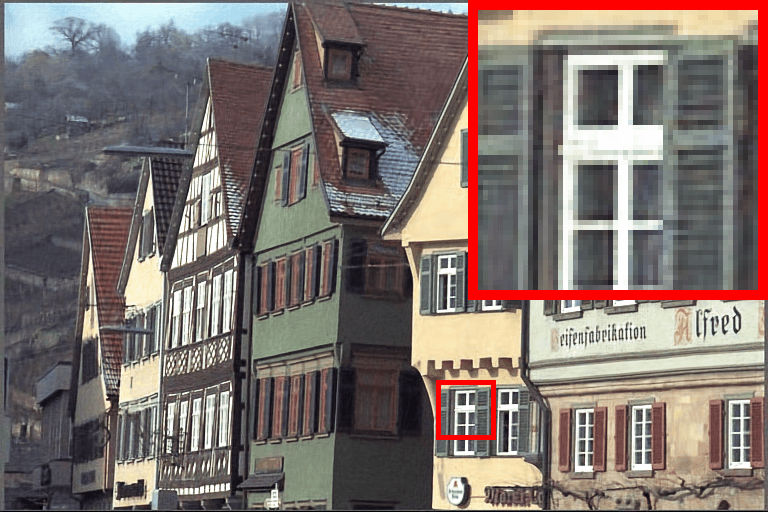} &
\includegraphics[width=0.156\linewidth]{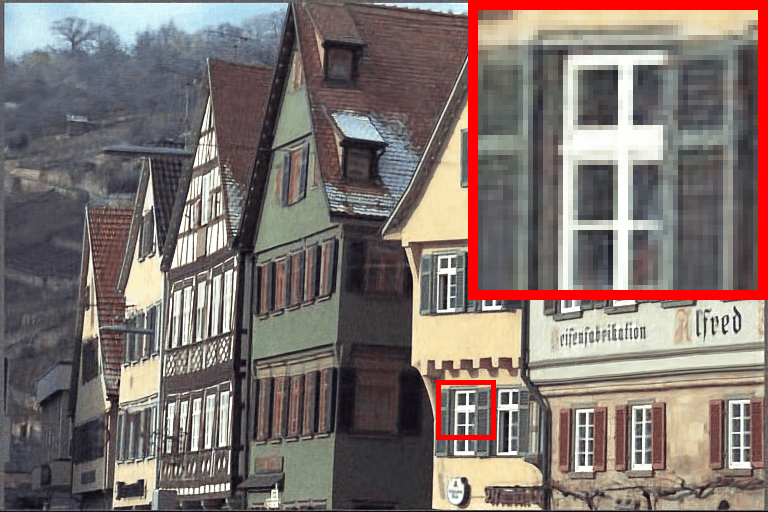} &
\includegraphics[width=0.156\linewidth]{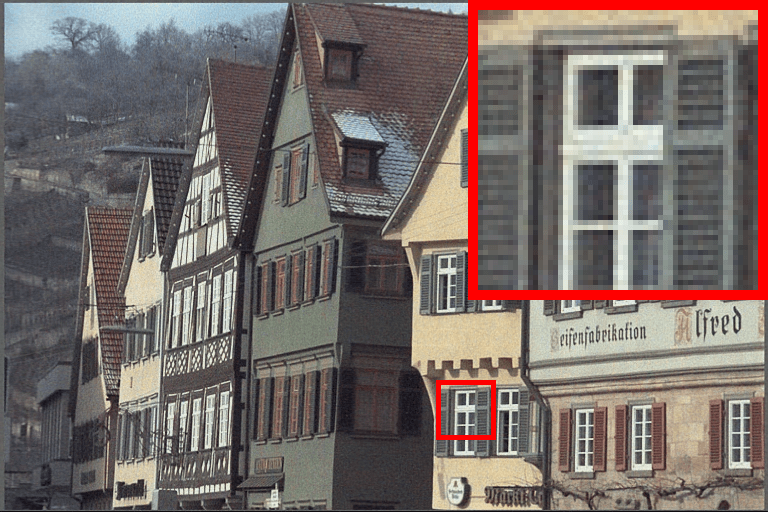} &
\includegraphics[width=0.156\linewidth]{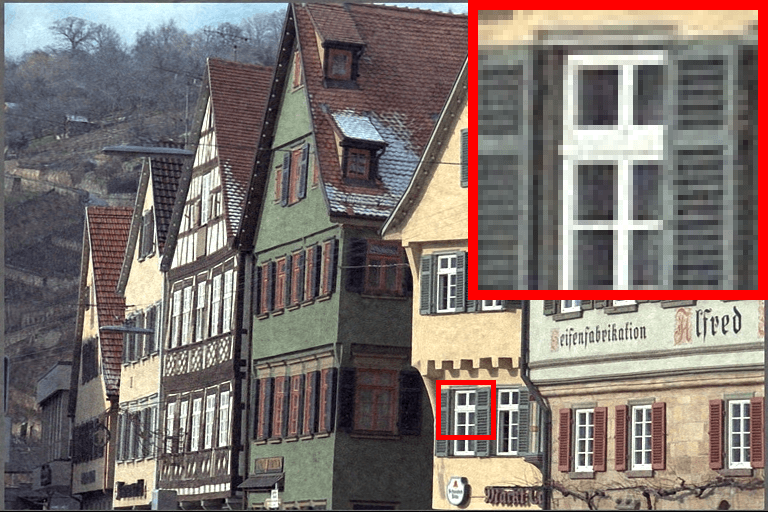} &
\includegraphics[width=0.156\linewidth]{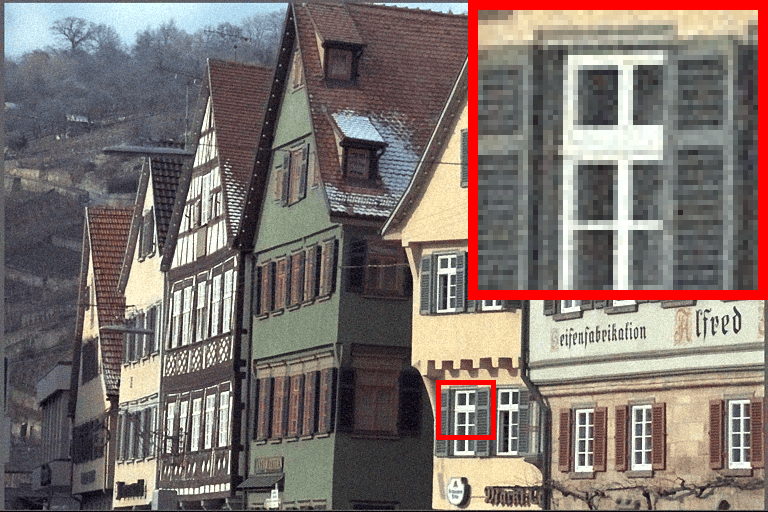}\\
Kodak-20  & 47.75dB & 35.74dB & 32.58dB & 39.00dB & 33.89dB & 37.06dB\\
\includegraphics[width=0.156\linewidth]{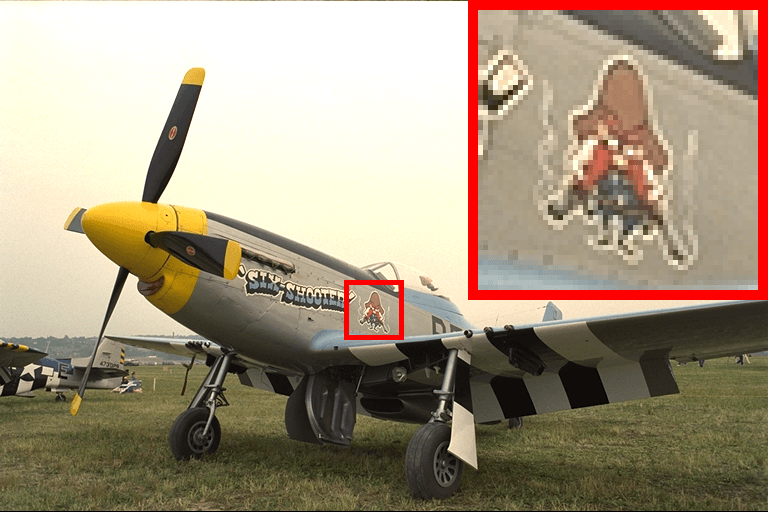} & 
\includegraphics[width=0.156\linewidth]{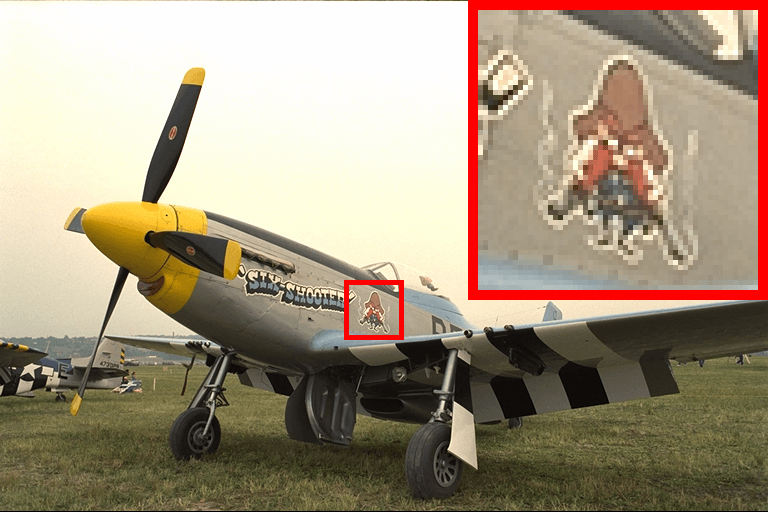} &
\includegraphics[width=0.156\linewidth]{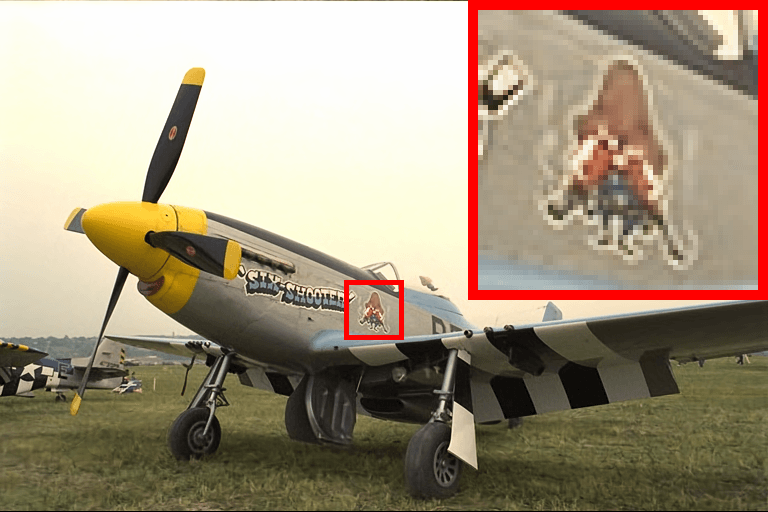} &
\includegraphics[width=0.156\linewidth]{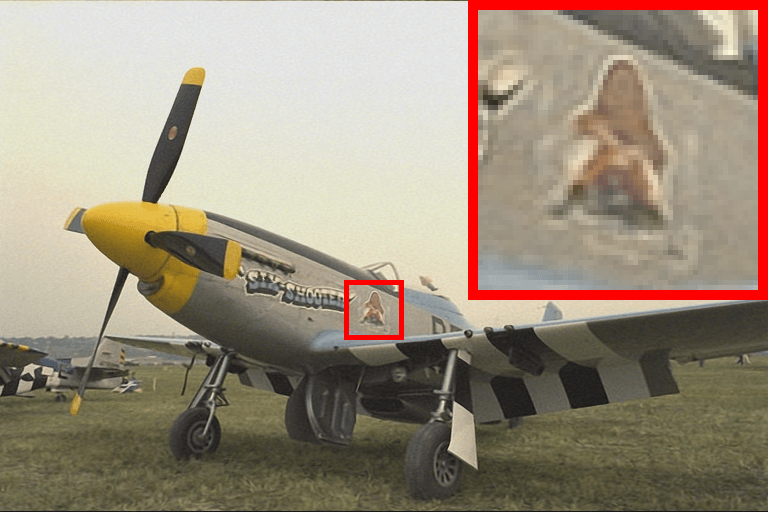} &
\includegraphics[width=0.156\linewidth]{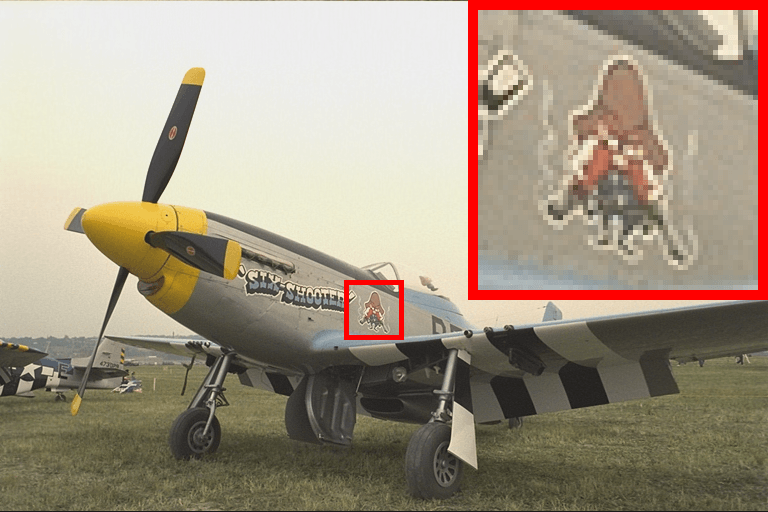} &
\includegraphics[width=0.156\linewidth]{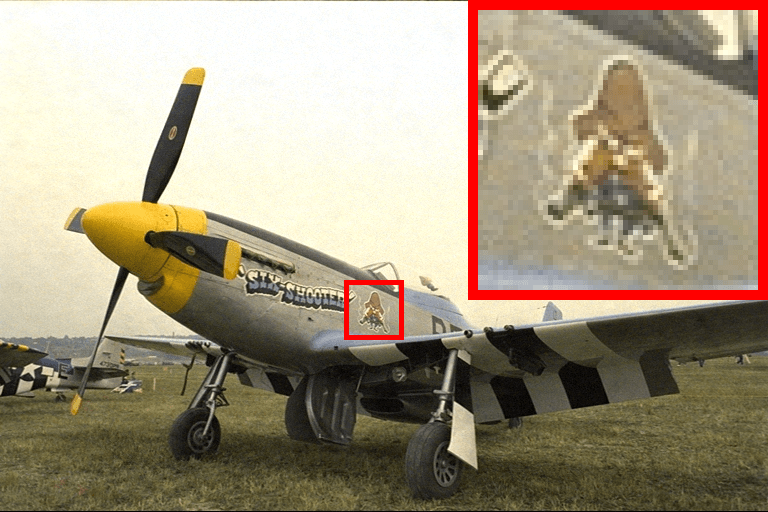} &
\includegraphics[width=0.156\linewidth]{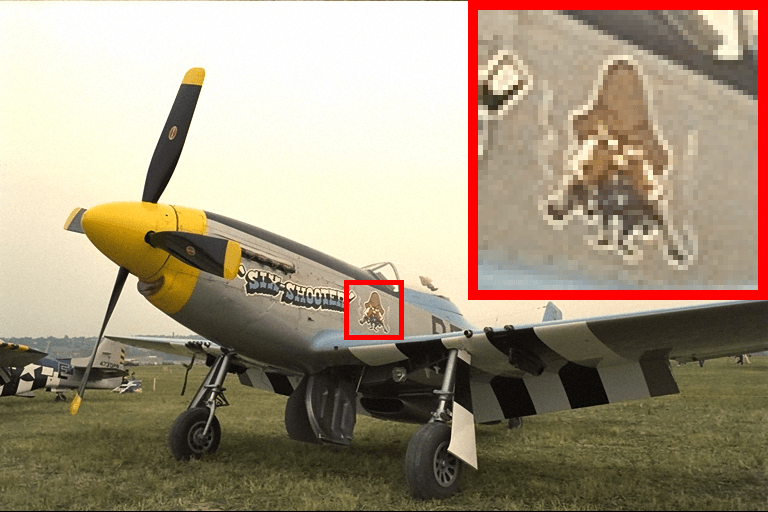}\\
Set14-11  & 32.47dB & 29.64dB & 29.52dB & 29.83dB & 27.09dB & 28.33dB\\
\includegraphics[width=0.156\linewidth]{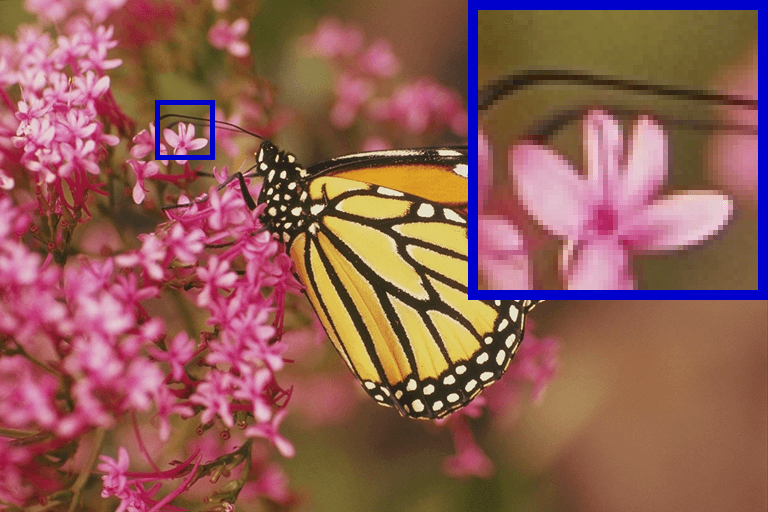} & 
\includegraphics[width=0.156\linewidth]{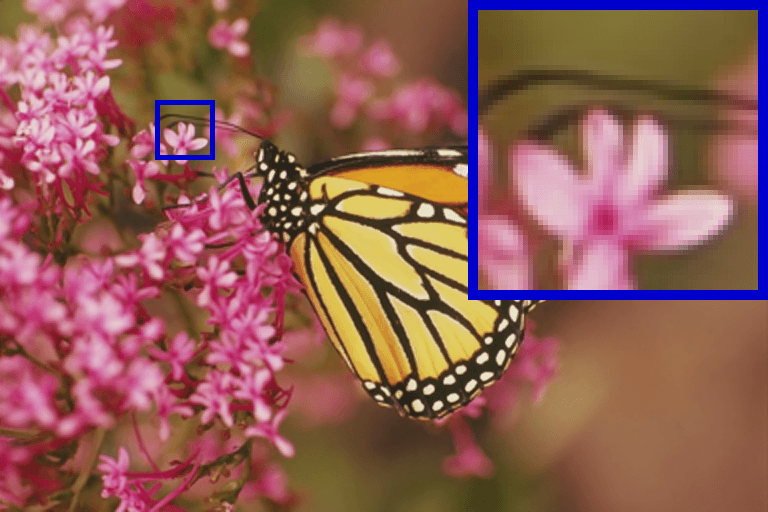} &
\includegraphics[width=0.156\linewidth]{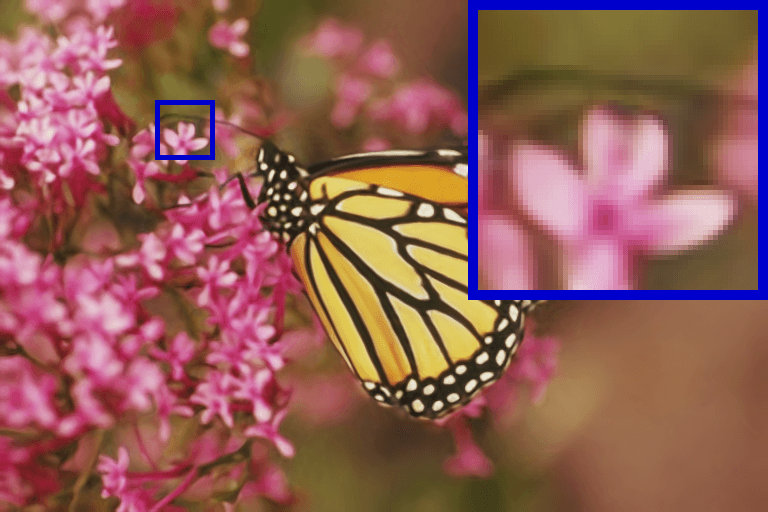} &
\includegraphics[width=0.156\linewidth]{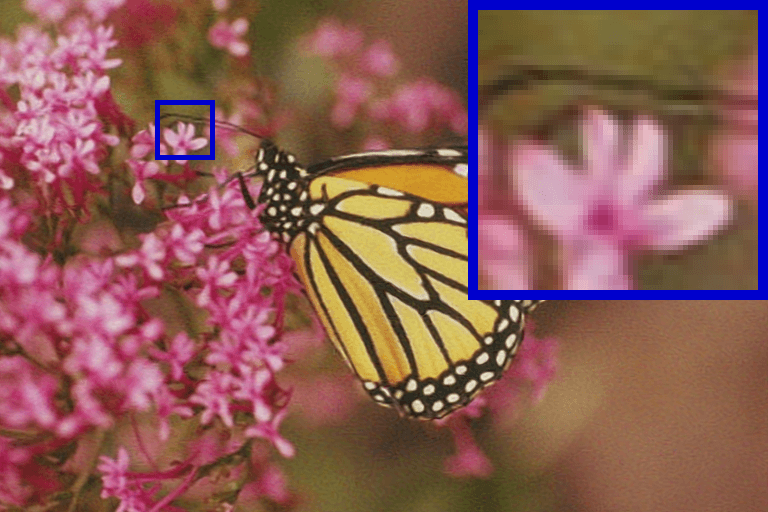} &
\includegraphics[width=0.156\linewidth]{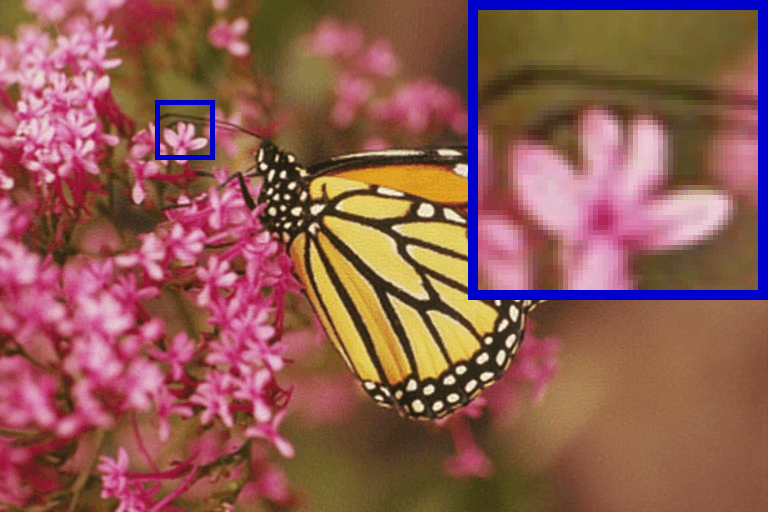} &
\includegraphics[width=0.156\linewidth]{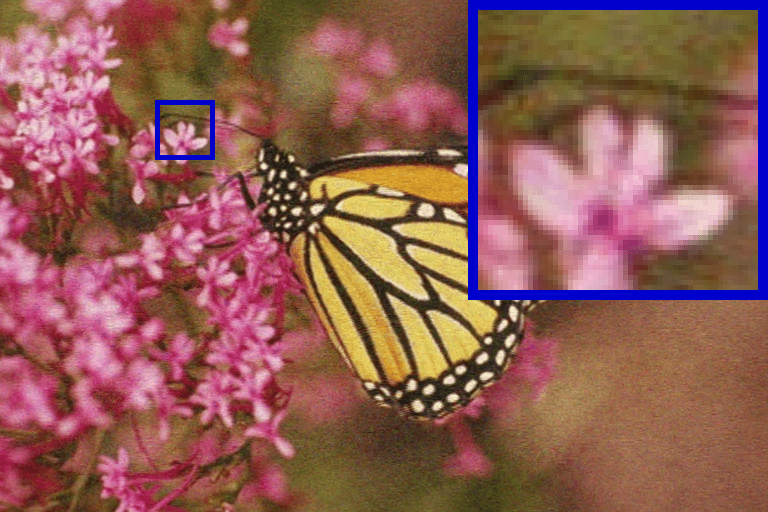} &
\includegraphics[width=0.156\linewidth]{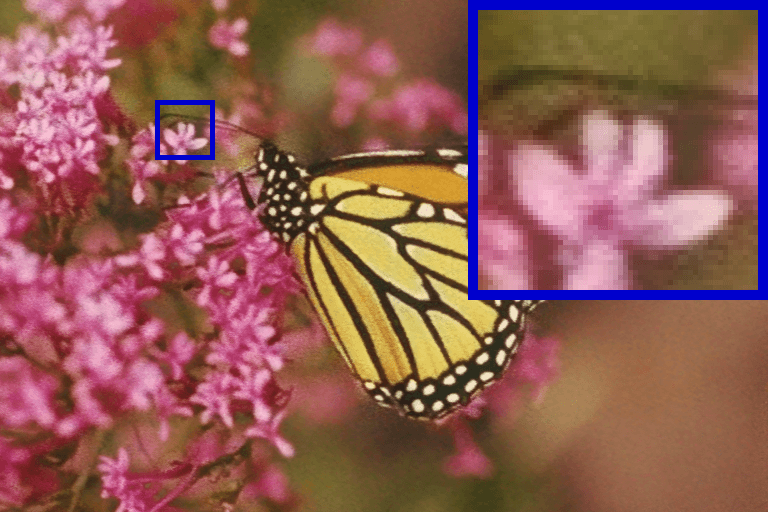}\\
Set14-14  & 29.48dB & 26.35dB & 26.17dB & 27.49dB & 25.51dB & 25.10dB\\
\includegraphics[width=0.156\linewidth]{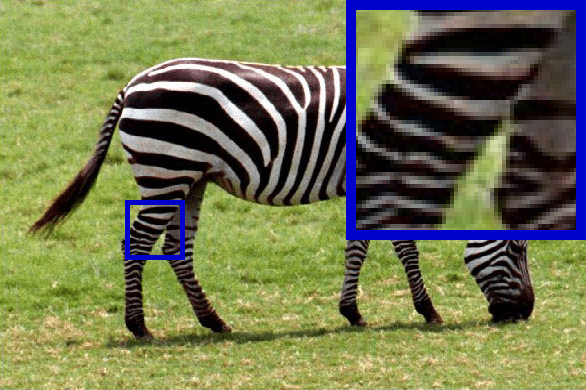} & 
\includegraphics[width=0.156\linewidth]{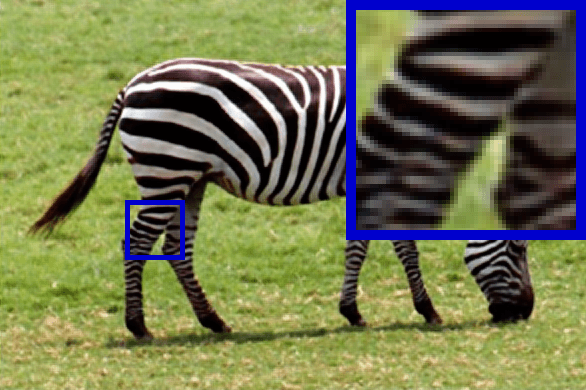} &
\includegraphics[width=0.156\linewidth]{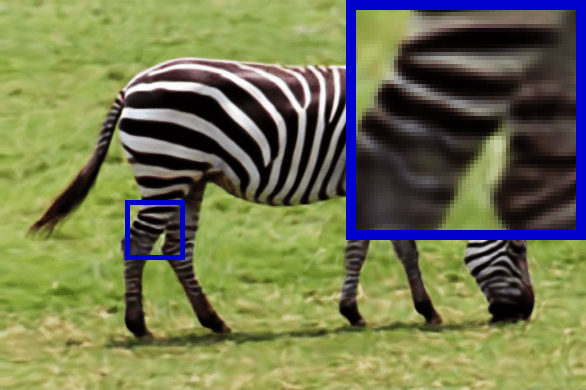} &
\includegraphics[width=0.156\linewidth]{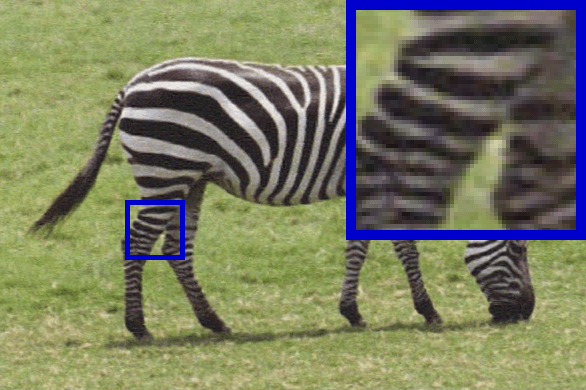} &
\includegraphics[width=0.156\linewidth]{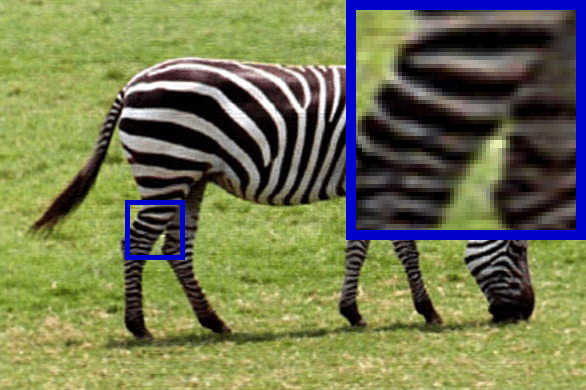} &
\includegraphics[width=0.156\linewidth]{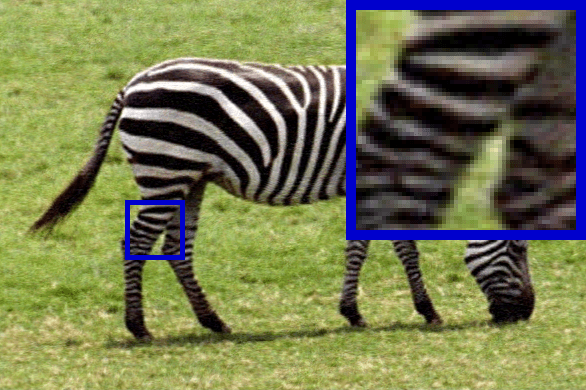} &
\includegraphics[width=0.156\linewidth]{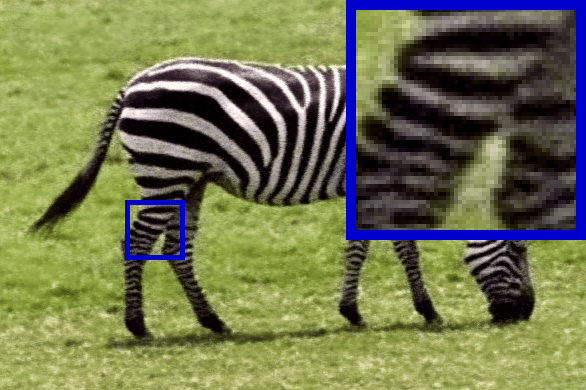}\\
\textbf{GT} & \textbf{C$^{2}$-INR (Ours)}& \textbf{DD} & \textbf{INCODE} & \textbf{SL$^{2}$A-INR} & \textbf{MIRE} & \textbf{MSA}\\
\end{tabular}
}
\caption{Qualitative comparisons on the Kodak~\cite{kodakdataset} and Set14~\cite{zeyde2010single} datasets. The first two rows present results for the image representation task, while the 3rd and 4th rows show results for the image $\times 2$ SR task. \js{Under identical parameter budget, our C$^{2}$-INR consistently produces cleaner images, with noticeable reductions in distortions and artifacts, e.g., the horizontal window mullions and aircraft fuselage paintings in the figures.}}
\label{fig:re_img}
\end{figure*}

\begin{figure}[!t]
    \centering
    \setlength{
    \tabcolsep}{1pt}
    \centering
\resizebox{0.49\textwidth}{!}{
\begin{tabular}{ccc}
\footnotesize
\textbf{GT}-Set5-03  & 
\footnotesize\textbf{Ours}-39.06dB & \footnotesize\textbf{DD}-36.46dB \\
\includegraphics[width=0.33\linewidth]{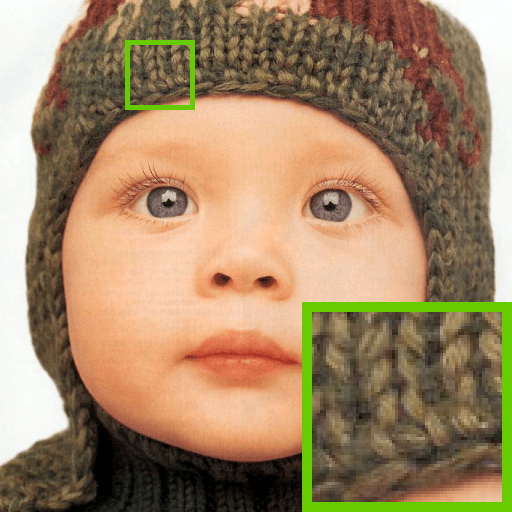} & 
\includegraphics[width=0.33\linewidth]{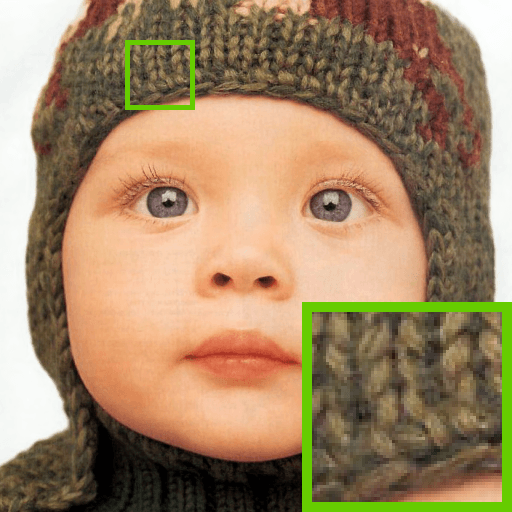} &
\includegraphics[width=0.33\linewidth]{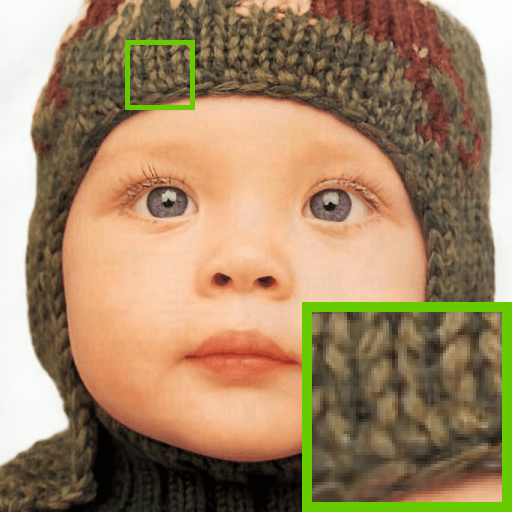}\\
\footnotesize\textbf{SL$^{2}$A-INR}-36.28dB  & \footnotesize\textbf{MIRE}-36.93dB & \footnotesize\textbf{MSA}-36.30dB \\
\includegraphics[width=0.33\linewidth]{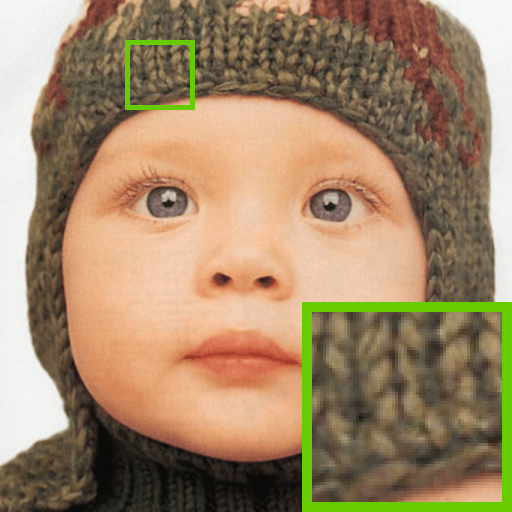} & 
\includegraphics[width=0.33\linewidth]{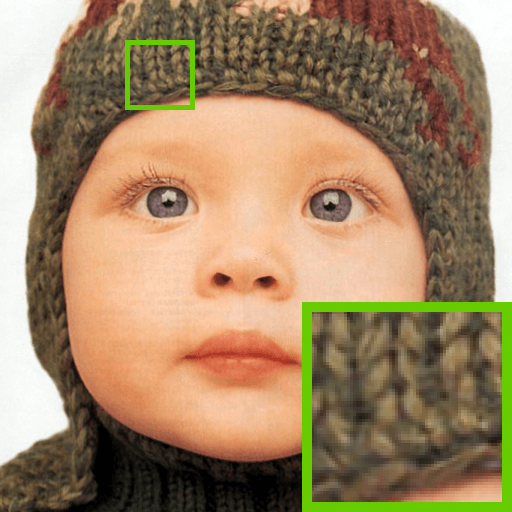} &
\includegraphics[width=0.33\linewidth]{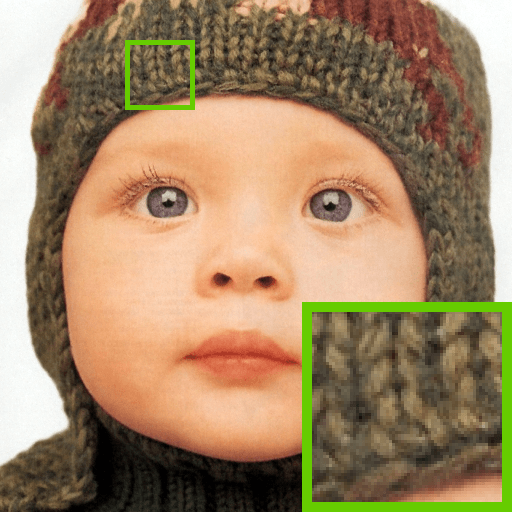}\\
\end{tabular}
}
\caption{Qualitative comparison on the Set5~\cite{bevilacqua2012low} dataset for the task of image inpainting (masking 30\% pixels). \js{Our C$^{2}$-INR more effectively leverages inter-pixel correlations to fill missing pixels, preserving the sharpness of the hat's woven texture and achieving superior visual quality.}}
\label{fig:in_img}
\end{figure}

\subsubsection{Image Super-resolution}
\jshi{We further evaluate C$^{2}$-INR against other SOTA frameworks on the image super‑resolution task. Using all 14 images from the Set14 dataset~\cite{zeyde2010single} as our test set, which provides both full‑resolution and $\times2$ downsampled versions, we apply bicubic interpolation to further downsample them for $\times4$ super‑resolution evaluation. All methods are compared under a parameter budget of 215k to ensure fairness. Table~\ref{tab:image_sr} reports the averaged PSNR and SSIM for both $\times2$ and $\times4$ tasks, along with the LPIPS metric to assess perceptual quality. The third and fourth rows of Fig.~\ref{fig:re_img} visualize the super‑resolved outputs from each method. Both DD and C$^{2}$-INR outperform the MLP‑based INR methods, confirming that convolution‑based architectures better exploit local pixel correlations to produce sharper details with fewer artifacts. Moreover, thanks to its image‑specific customized design, C$^{2}$-INR achieves superior performance over DD.
}

\subsubsection{Image Inpainting}
\jshi{
Following SL$^{2}$A-INR~\cite{heidari2024sl2a}, we further evaluate each method on the more challenging task of image inpainting. Specifically, all methods are tested under the same parameter budget (215k parameters) using all 5 images from the Set5 dataset~\cite{bevilacqua2012low}. We generate two random masks covering 30\% and 50\% of the image pixels, respectively, and adopt each method to recover the masked regions from the known pixels. Table~\ref{tab:image_inpainting} reports the averaged PSNR, SSIM, and LPIPS for each method, and Fig.~\ref{fig:in_img} visualizes the inpainted results. As shown, C$^{2}$-INR produces images with more realistic details and finer textures than the other methods, demonstrating its clear superiority over existing SOTA approaches.
}

\subsection{Ablation Studies}
\subsubsection{Module Design Effectiveness}
\jshi{
To systematically investigate the contribution of each module design to the overall performance of C$^{2}$-INR, we conduct a step-by-step ablation study on four key modules, i.e. Directional Kernel Allocation (DKA), Kernel Reformulation with Basis (KRB), Activation Function Selection (AFS), and Progressive Layer Fitting (PLF). Using a cascaded convolutional network as the backbone, we keep the total number of parameters fixed at 215k and incrementally integrate each module into the backbone. All variants are evaluated on the Kodak dataset~\cite{kodakdataset} in terms of averaged performance for image representation tasks. As reported in Table~\ref{tab:modular_impact}, the progressive integration of these modules leads to consistent improvements in the performance of C$^{2}$-INR. Specifically, the introduction of DKA and KRB jointly boosts the PSNR from 39.60dB to 42.60dB, yielding a gain of 3dB. Subsequently, adopting AFS and PLF improves the PSNR by 1.9dB, finally reaching 44.51dB. These results demonstrate that tailoring the two core components, the convolutional kernels and the activation functions, can significantly enhance the representation capacity of convolutional INR, thereby validating the effectiveness and rationality of our proposed module designs.
}

\begin{table}[t]
	\tiny
	\setlength\tabcolsep{2pt}
	\centering
\resizebox{0.45\textwidth}{!}{
\begin{tabular}{l|cccc|cc}
\toprule
    Variants& DKA& KRB& AFS& PLF & PSNR$\uparrow$ & SSIM $\uparrow$
    \\ 
    \hline
    C$^{2}$-INR*  & \textcolor{gray}{\XSolidBrush} & \textcolor{gray}{\XSolidBrush} & \textcolor{gray}{\XSolidBrush}  &  \textcolor{gray}{\XSolidBrush} & 39.60 & 0.9634 \\
    C$^{2}$-INR$^{\dagger}$  & \textcolor{black}{\Checkmark} & \textcolor{gray}{\XSolidBrush} & \textcolor{gray}{\XSolidBrush}  &  \textcolor{gray}{\XSolidBrush} & 41.39 & 0.9736 \\
    C$^{2}$-INR$^{\ddagger}$  & \textcolor{black}{\Checkmark} & \textcolor{black}{\Checkmark} & \textcolor{gray}{\XSolidBrush}  &  \textcolor{gray}{\XSolidBrush} & 42.60 & 0.9789 \\ 
    C$^{2}$-INR$^{\star}$  & \textcolor{black}{\Checkmark} & \textcolor{black}{\Checkmark} & \textcolor{black}{\Checkmark}  &  \textcolor{gray}{\XSolidBrush} & 44.03 & 0.9831 \\ 
    \hline
     C$^{2}$-INR    &
     \textcolor{black}{\Checkmark} & \textcolor{black}{\Checkmark} & \textcolor{black}{\Checkmark}  &  
     \textcolor{black}{\Checkmark} & \textbf{44.51} & \textbf{0.9846}  \\
    \toprule
\end{tabular}}
\caption{Ablation study on different modules. 
\js{The baseline ($*$) denotes the vanilla C$^{2}$-INR without any of our enhancements. The variants $\dagger$, $\ddagger$, and $\star$ correspond to progressive integrations of DKA, KRB, and AFS, respectively, with C$^{2}$-INR representing the full model.}
}
\label{tab:modular_impact}
\end{table}

\begin{figure}[t]
    \centering
    \setlength{
    \tabcolsep}{1pt}
    \centering
\resizebox{0.49\textwidth}{!}{
\begin{tabular}{c@{\hspace{0.01cm}}c}
\includegraphics[width=0.27\linewidth,height=0.16\linewidth]{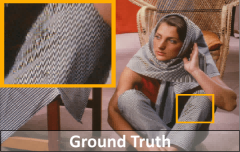} & 
\includegraphics[width=0.27\linewidth,height=0.16\linewidth]{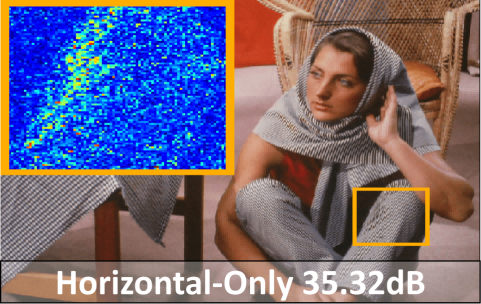} \\[-0.11cm]
\includegraphics[width=0.27\linewidth,height=0.16\linewidth]{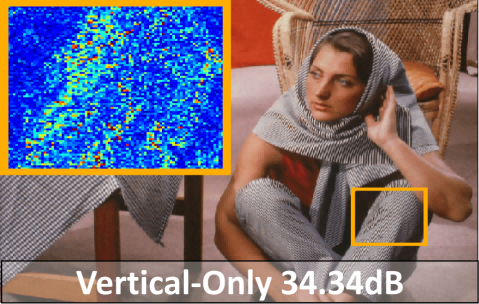} & 
\includegraphics[width=0.27\linewidth,height=0.16\linewidth]{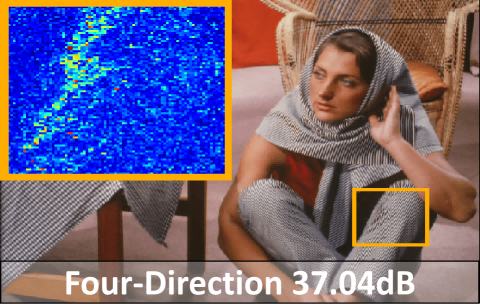} \\
\end{tabular}
}
\caption{\js{Image and error map visualizations for a horizontally dominated example (108k params), comparing horizontal-only, vertical-only, and four-direction kernel settings, suggesting that kernel allocation should adapt to image content.}}
\label{fig:error_map}
\end{figure}

\subsubsection{Directional Kernel Allocation}
\jshi{
\js{Besides the modular ablation, we further evaluate several allocation strategies: standard $3\times 3$ kernels, two-direction sets ($0^{\circ}$\&$90^{\circ}$ and $45^{\circ}$\&$135^{\circ}$), and our four-direction set, all with equal proportions. We also compare two content‑adaptive criteria: gradient‑based (local dominant direction) and spectral‑energy‑based (FT). Experiments on Kodak~\cite{kodakdataset} for image representation are summarized in Table~\ref{tab:kernel_allocation}.
Results show that directional kernels consistently outperform standard $3\times 3$ under the same parameter budget, with the four-direction set surpassing two-direction ones. Gradient‑based allocation, relying only on local first‑order statistics, performs even worse than uniform allocation. In contrast, our spectral criterion accurately measures directional parameter demands, enabling content‑adaptive customization and achieving the best performance. This advantage is further visualized in Fig.~\ref{fig:error_map} on a horizontally dominated image (spectrum in Fig.~\ref{fig:example_spectrum}, bottom). Compared with horizontal-only and vertical-only kernels, our four-direction setting yields noticeably fewer artifacts (particularly along the horizontal scarf textures), demonstrating the effectiveness of our allocation strategy.}
}

\subsubsection{Activation Function Selection Strategy}
\begin{figure}[t]
    \centering
\includegraphics[width=0.48\textwidth,height=0.18\textwidth]{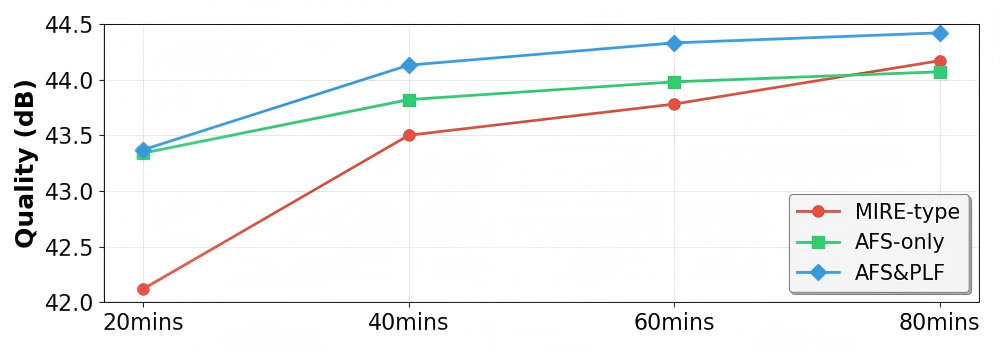}
    \caption{Representation quality evolution curves for each activation function selection strategies, implying that our joint AFS and PLF stategies gives the best efficiency.}
    \label{fig:rep_curves}
\end{figure}

\jshi{
To evaluate the effectiveness of our activation function selection strategy, we compare three different schemes: (1) MIRE-type, which fits each activation function layer by layer and selects the single best one for each layer based on reconstruction quality; (2) AFS-only, which adopts the annealed activation function selection mechanism without progressive layer fitting, i.e., simultaneously selecting the optimal activation function for all channels across all layers; and (3) AFS\&PLF, our proposed strategy that combines annealed activation function selection with progressive layer fitting. We compare the network representation performance under different selection strategies at various training time points, with results shown in Fig.~\ref{fig:rep_curves}. As observed, the MIRE-type strategy selects the best function per layer, but its granularity is at the layer level rather than the kernel level, leading to lower training efficiency and performance. The AFS-only strategy achieves kernel‑level granularity, yet optimizing all layers simultaneously makes it more prone to falling into local minima. In contrast, our AFS\&PLF strikes a favorable balance between granularity and convergence efficiency, thereby achieving the best representation performance within a given training time.
}

\begin{table}[t!]
	\tiny
	\setlength\tabcolsep{2pt}
	\centering
\resizebox{0.44\textwidth}{!}{
\begin{tabular}{l|c|cc}
\toprule
    Kernel Types& Proportions & PSNR$\uparrow$ & SSIM $\uparrow$
    \\ 
    \hline
     $3\times3$ & Equal Prop. & 42.76 & 0.9788 \\
    $0^{\circ}$+$90^{\circ}$ & Equal Prop. & 43.79 & 0.9835\\
    $45^{\circ}$+$135^{\circ}$ & Equal Prop. & 44.03 & 0.9840\\
    $0^{\circ}$+$90^{\circ}$+$45^{\circ}$+$135^{\circ}$ & Equal Prop. & 44.30 & 0.9843\\
   $0^{\circ}$+$90^{\circ}$+$45^{\circ}$+$135^{\circ}$ & w.r.t. Grad. & 44.20 & 0.9841\\ 
    \hline
    $0^{\circ}$+$90^{\circ}$+$45^{\circ}$+$135^{\circ}$ & w.r.t. FT & \textbf{44.51} & \textbf{0.9846}
  \\
    \toprule
\end{tabular}}
\caption{Ablation on different kernel allocation strategies.}
\label{tab:kernel_allocation}
\end{table}

\section{Conclusion}
\js{
We propose C$^{2}$-INR, a customized C-INR that adapts both its convolution kernels and activation functions to the target image. For kernels, we introduce directional filters with proportions adaptively allocated according to spectral energy, assigning more capacity to dominant directions. For activations functions, we select the most suitable function per kernel via response‑based scoring with an annealing mechanism, and adopt progressive layer‑wise training for stable convergence. Extensive experiments on representation, SR, and inpainting show that C$^{2}$-INR consistently outperforms SOTA methods. While currently 2D‑specific and memory‑intensive for high resolution, future work will extend it to higher dimensions and incorporate partition strategies to reduce memory footprint.
}

\bibliography{aaai2027}

@String(CVPR= {IEEE Conf. Comput. Vis. Pattern Recog. (CVPR)})

@String(ICCV= {IEEE Int. Conf. Comput. Vis. (ICCV)})

@String(ECCV= {Eur. Conf. Comput. Vis. (ECCV)})

@String(NeurIPS= {Adv. Neural Inform. Process. Syst. (NeurIPS)})

@String(WACV= {IEEE Winter Conf. Appl. Comput. Vis. (WACV)})

@String(BMVC= {Brit. Mach. Vis. Conf. (BMVC)})

@String(TIP  = {IEEE Trans. Image Process. (TIP)})

@String(ICLR = {Int. Conf. Learn. Represent. (ICLR)})

@String(AAAI = {AAAI Conf. Artif. Intell. (AAAI)})

@String(ICML = {Int. Conf. on Mach. Learn. (ICML)})

@inproceedings{shi2024inductive,
  title={Inductive Gradient Adjustment For Spectral Bias In Implicit Neural Representations},
  author={Kexuan Shi and Hai Chen and Leheng Zhang and Shuhang Gu},
  booktitle=ICML,
  year={2025}
}

@inproceedings{heidari2024sl2a,
  title={SL$^{2}$A-INR: Single-Layer Learnable Activation for Implicit Neural Representation},
  author={Heidari, Moein and Rezaeian, Reza and Azad, Reza and Merhof, Dorit and Soltanian-Zadeh, Hamid and Hacihaliloglu, Ilker},
  booktitle=ICCV,
  year={2025}
}

@inproceedings{haider2026inr,
  title={I-{INR}: iterative implicit neural representations},
  author={Haider, Ali and Ali, Muhammad Salman and Qamar, Maryam and Khalil, Tahir and Kim, Soo Ye and Oh, Jihyong and Tartaglione, Enzo and Bae, Sung-Ho},
  booktitle=AAAI,
  pages={4520--4528},
  year={2026}
}

@inproceedings{zhao2025adaptive,
  title={Adaptive wavelet-positional encoding for high-frequency information learning in implicit neural representation},
  author={Zhao, Hongxu and Gao, Zelin and Wang, Yue and Xiong, Rong and Zhang, Yu},
  booktitle=AAAI,
  pages={10430--10438},
  year={2025}
}

@inproceedings{li2023regularize,
  title={Regularize implicit neural representation by itself},
  author={Li, Zhemin and Wang, Hongxia and Meng, Deyu},
  booktitle=CVPR,
  pages={10280--10288},
  year={2023}
}

@inproceedings{jayasundara2025mire,
  title={Mire: Matched implicit neural representations},
  author={Jayasundara, Dhananjaya and Zhao, Heng and Labate, Demetrio and Patel, Vishal M},
  booktitle=CVPR,
  pages={8279--8288},
  year={2025}
}

@inproceedings{han2026implicit,
  title={Implicit Neural Representation with Multi-Scale Sine Activation},
  author={Han, Jufeng and Wei, Shu and Wu, Min and Yu, Lina and Li, Weijun and Sun, Linjun and Qin, Hong and Pang, Yan},
  booktitle=AAAI,
  pages={21567--21575},
  year={2026}
}

@article{zhang2026understanding,
  title={Understanding bias terms in neural representations},
  author={Zhang, Weixiang and Li, Boxi and Xie, Shuzhao and Ren, Chengwei and Xue, Yuan and Wang, Zhi},
  journal=NeurIPS,
  pages={103682--103705},
  year={2026}
}

@inproceedings{heckel2018deep,
  title={Deep decoder: Concise image representations from untrained non-convolutional networks},
  author={Heckel, Reinhard and Hand, Paul},
  booktitle=ICLR,
  year={2019}
}

@article{chen2021nerv,
  title={Nerv: Neural representations for videos},
  author={Chen, Hao and He, Bo and Wang, Hanyu and Ren, Yixuan and Lim, Ser Nam and Shrivastava, Abhinav},
  journal=NeurIPS,
  pages={21557--21568},
  year={2021}
}

@inproceedings{chen2023hnerv,
  title={Hnerv: A hybrid neural representation for videos},
  author={Chen, Hao and Gwilliam, Matthew and Lim, Ser-Nam and Shrivastava, Abhinav},
  booktitle=CVPR,
  pages={10270--10279},
  year={2023}
}

@inproceedings{zhang2024boosting,
  title={Boosting neural representations for videos with a conditional decoder},
  author={Zhang, Xinjie and Yang, Ren and He, Dailan and Ge, Xingtong and Xu, Tongda and Wang, Yan and Qin, Hongwei and Zhang, Jun},
  booktitle=CVPR,
  pages={2556--2566},
  year={2024}
}

@article{kwan2024immersive,
  title={Immersive video compression using implicit neural representations},
  author={Kwan, Ho Man and Zhang, Fan and Gower, Andrew and Bull, David},
  journal={Picture Coding Symposium (PCS)},
  year={2024}
}

@article{shi2024learning,
  title={Learning kernel-modulated neural representation for efficient light field compression},
  author={Shi, Jinglei and Xu, Yihong and Guillemot, Christine},
  journal=TIP,
  volume={33},
  pages={4060--4074},
  year={2024}
}

@article{su2022inras,
  title={Inras: Implicit neural representation for audio scenes},
  author={Su, Kun and Chen, Mingfei and Shlizerman, Eli},
  journal=NeurIPS,
  pages={8144--8158},
  year={2022}
}

@inproceedings{ramasinghe2022beyond,
  title={Beyond periodicity: Towards a unifying framework for activations in coordinate-mlps},
  author={Ramasinghe, Sameera and Lucey, Simon},
  booktitle=ECCV,
  pages={142--158},
  year={2022}
}

@inproceedings{saragadam2023wire,
  title={Wire: Wavelet implicit neural representations},
  author={Saragadam, Vishwanath and LeJeune, Daniel and Tan, Jasper and Balakrishnan, Guha and Veeraraghavan, Ashok and Baraniuk, Richard G},
  booktitle=CVPR,
  pages={18507--18516},
  year={2023}
}

@inproceedings{jayasundara2025pin,
  title={PIN: Prolate spheroidal wave function-based implicit neural representations},
  author={Jayasundara Mudiyanselage, Viraj Dhananjaya Bandara Jayasundara and Zhao, Heng and Labate, Demetrio and Patel, Vishal},
  booktitle=ICLR,
  pages={17832--17853},
  year={2025}
}

@inproceedings{yan2024ds,
  title={Ds-nerv: Implicit neural video representation with decomposed static and dynamic codes},
  author={Yan, Hao and Ke, Zhihui and Zhou, Xiaobo and Qiu, Tie and Shi, Xidong and Jiang, Dadong},
  booktitle=CVPR,
  pages={23019--23029},
  year={2024}
}

@inproceedings{gao2025givic,
  title={GIViC: Generative implicit video compression},
  author={Gao, Ge and Teng, Siyue and Peng, Tianhao and Zhang, Fan and Bull, David},
  booktitle=ICCV,
  pages={17356--17367},
  year={2025}
}

@inproceedings{guo2025metanerv,
  title={Metanerv: Meta neural representations for videos with spatial-temporal guidance},
  author={Guo, Jialong and Liu, Ke and Yao, Jiangchao and Wang, Zhihua and Bu, Jiajun and Wang, Haishuai},
  booktitle=AAAI,
  volume={39},
  pages={3257--3265},
  year={2025}
}

@inproceedings{wu2024tetrirf,
  title={Tetrirf: Temporal tri-plane radiance fields for efficient free-viewpoint video},
  author={Wu, Minye and Wang, Zehao and Kouros, Georgios and Tuytelaars, Tinne},
  booktitle=CVPR,
  pages={6487--6496},
  year={2024}
}

@article{zhu2025implicit,
  title={Implicit-explicit integrated representations for multi-view video compression},
  author={Zhu, Chen and Lu, Guo and He, Bing and Xie, Rong and Song, Li},
  journal=TIP,
  volume={34},
  pages={1106--1118},
  year={2025}
}

@article{sitzmann2020implicit,
  title={Implicit neural representations with periodic activation functions},
  author={Sitzmann, Vincent and Martel, Julien and Bergman, Alexander and Lindell, David and Wetzstein, Gordon},
  journal=NeurIPS,
  volume={33},
  pages={7462--7473},
  year={2020}
}

@inproceedings{liu2024finer,
  title={Finer: Flexible spectral-bias tuning in implicit neural representation by variable-periodic activation functions},
  author={Liu, Zhen and Zhu, Hao and Zhang, Qi and Fu, Jingde and Deng, Weibing and Ma, Zhan and Guo, Yanwen and Cao, Xun},
  booktitle=CVPR,
  pages={2713--2722},
  year={2024}
}

@article{liu2020multi,
  title={Multi-scale deep neural network ({MscaleDNN}) for solving Poisson-Boltzmann equation in complex domains},
  author={Liu, Ziqi and Cai, Wei and Xu, Zhi-Qin John},
  journal={Commun. Comput. Phys. (CCP)},
  year={2020}
}

@article{zhu2025msnerv,
  title={MSNeRV: neural video representation with multi-scale feature fusion},
  author={Zhu, Jun and Zhang, Xinfeng and Tang, Lv and Jiang, JunHao},
  journal={arXiv preprint arXiv:2506.15276},
  year={2025}
}

@inproceedings{jiang2025hiif,
  title={{HIIF}: Hierarchical encoding based implicit image function for continuous super-resolution},
  author={Jiang, Yuxuan and Kwan, Ho Man and Peng, Tianhao and Gao, Ge and Zhang, Fan and Zhu, Xiaoqing and Sole, Joel and Bull, David},
  booktitle=CVPR,
  pages={2289--2299},
  year={2025}
}

@inproceedings{zhao2024pnerv,
  title={Pnerv: Enhancing spatial consistency via pyramidal neural representation for videos},
  author={Zhao, Qi and Asif, M Salman and Ma, Zhan},
  booktitle=CVPR,
  pages={19103--19112},
  year={2024}
}

@article{mildenhall2021nerf,
  title={{NeRF}: Representing scenes as neural radiance fields for view synthesis},
  author={Mildenhall, Ben and Srinivasan, Pratul P and Tancik, Matthew and Barron, Jonathan T and Ramamoorthi, Ravi and Ng, Ren},
  journal={Communications of the ACM},
  volume={65},
  number={1},
  pages={99--106},
  year={2021}
}

@article{ashkenazi2024towards,
  title={Towards croppable implicit neural representations},
  author={Ashkenazi, Maor and Treister, Eran},
  journal=NeurIPS,
  volume={37},
  pages={31473--31503},
  year={2024}
}

@article{tang2025canerv,
  title={Canerv: Content adaptive neural representation for video compression},
  author={Tang, Lv and Zhu, Jun and Zhang, Xinfeng and Zhang, Li and Ma, Siwei and Huang, Qingming},
  journal={arXiv preprint arXiv:2502.06181},
  year={2025}
}

@inproceedings{kazerouni2024incode,
  title={Incode: Implicit neural conditioning with prior knowledge embeddings},
  author={Kazerouni, Amirhossein and Azad, Reza and Hosseini, Alireza and Merhof, Dorit and Bagci, Ulas},
  booktitle=WACV,
  pages={1298--1307},
  year={2024}
}

@article{goyal2019learning,
  title={Learning activation functions: A new paradigm for understanding neural networks},
  author={Goyal, Mohit and Goyal, Rajan and Lall, Brejesh},
  journal={arXiv preprint arXiv:1906.09529},
  year={2019}
}

@article{bingham2022discovering,
  title={Discovering parametric activation functions},
  author={Bingham, Garrett and Miikkulainen, Risto},
  journal={Neural Networks (NN)},
  volume={148},
  pages={48--65},
  year={2022}
}

@article{fakhoury2022exsplinet,
  title={ExSpliNet: An interpretable and expressive spline-based neural network},
  author={Fakhoury, Daniele and Fakhoury, Emanuele and Speleers, Hendrik},
  journal={Neural Networks (NN)},
  volume={152},
  pages={332--346},
  year={2022}
}

@inproceedings{zhou2017oriented,
  title={Oriented response networks},
  author={Zhou, Yanzhao and Ye, Qixiang and Qiu, Qiang and Jiao, Jianbin},
  booktitle=CVPR,
  pages={519--528},
  year={2017}
}

@inproceedings{weiler2018learning,
  title={Learning steerable filters for rotation equivariant cnns},
  author={Weiler, Maurice and Hamprecht, Fred A and Storath, Martin},
  booktitle=CVPR,
  pages={849--858},
  year={2018}
}

@inproceedings{srinivasan2021nerv,
title={Nerv: Neural reflectance and visibility fields for relighting and view synthesis},
author={Srinivasan, Pratul P and Deng, Boyang and Zhang, Xiuming and Tancik, Matthew and Mildenhall, Ben and Barron, Jonathan T},
booktitle=CVPR,
year={2021}
}

@inproceedings{park2019deepsdf,
  title={Deepsdf: Learning continuous signed distance functions for shape representation},
  author={Park, Jeong Joon and Florence, Peter and Straub, Julian and Newcombe, Richard and Lovegrove, Steven},
  booktitle=CVPR,
  pages={165--174},
  year={2019}
}

@article{wang2004image,
  title={Image quality assessment: from error visibility to structural similarity},
  author={Wang, Zhou and Bovik, Alan C and Sheikh, Hamid R and Simoncelli, Eero P},
  journal=TIP,
  volume={13},
  number={4},
  pages={600--612},
  year={2004}
}

@inproceedings{zhang2018unreasonable,
  title={The unreasonable effectiveness of deep features as a perceptual metric},
  author={Zhang, Richard and Isola, Phillip and Efros, Alexei A and Shechtman, Eli and Wang, Oliver},
  booktitle=CVPR,
  pages={586--595},
  year={2018}
}

@inproceedings{kodakdataset,
  title={Kodak lossless true color image suite},
  author={Kodak},
  year={2018}
}

@inproceedings{zeyde2010single,
  title={On single image scale-up using sparse-representations},
  author={Zeyde, Roman and Elad, Michael and Protter, Matan},
  booktitle={Int. Conf. on Curves and Surfaces (ICCS)},
  pages={711--730},
  year={2010}
}

@inproceedings{bevilacqua2012low,
  title={Low-complexity single-image super-resolution based on nonnegative neighbor embedding},
  author={Bevilacqua, Marco and Roumy, Aline and Guillemot, Christine and Morel, Marie-Line Alberi},
  booktitle=BMVC,
  year={2012}
}
\end{document}